\documentclass[graybox]{svmult}

\usepackage{mathptmx}       
\usepackage{helvet}         
\usepackage{courier}        
\usepackage{type1cm}        
\usepackage{makeidx}         
\usepackage{graphicx}        
\usepackage{multicol}        
\usepackage[bottom]{footmisc}

\usepackage{amsmath,amssymb,mathrsfs}
\usepackage[ruled, linesnumbered, vlined]{algorithm2e}
\usepackage{comment}
\usepackage{graphicx}
\usepackage{graphics}
\usepackage[tight]{subfigure}
\usepackage{wrapfig}
\usepackage{float}
\usepackage[table]{xcolor}
\usepackage{tabu}
\usepackage{multirow}
\usepackage{array}
\usepackage{enumerate}
\usepackage{sidecap}
\usepackage{flushend}
\usepackage{hyperref}
\usepackage{url}
\usepackage{booktabs}
\graphicspath{{./figs/}}

\usepackage[space]{cite}

\usepackage[shortlabels]{enumitem}
                    \setlist[enumerate, 1]{1\textsuperscript{o}}

\usepackage{color}

\definecolor{lightblue}{rgb}{0,0.2,1}
\definecolor{black}{rgb}{0,0,0}
\hypersetup{
    colorlinks=true,%
    citebordercolor=white,%
    filebordercolor=white,%
    linkbordercolor=white,%
    urlbordercolor=white,%
    citecolor=black,%
    linkcolor=black,%
    urlcolor=black%
}

\newcounter{tecounter}
\newenvironment{spmatrix}[1]
 {\def\mysubscript{#1}\mathop\bgroup\begin{pmatrix}}
 {\end{pmatrix}\egroup_{\textstyle\mathstrut\mysubscript}}

\makeindex             

\begin{document}

\title*{
Reachability and Differential based 
Heuristics for 
Solving Markov Decision Processes
}
\titlerunning{Reachability and Differential based Heuristics for Solving MDPs}
\author{Shoubhik Debnath$^{1}$, Lantao Liu$^{2}$, Gaurav Sukhatme$^{1}$}

\authorrunning{S. Debnath, L. Liu, G.S. Sukhatme} 
\institute{$^{1}$Shoubhik Debnath and Gaurav Sukhatme are with the Department of Computer Science at the University of Southern California, Los Angeles, CA 90089, USA. E-mail:
        {\tt\small \{sdebnath,gaurav\}@usc.edu \vspace{2pt}} \\
$^{2}$Lantao Liu is with the Intelligent Systems Engineering Department at  Indiana University,
        Bloomington, IN 47408, USA. E-mail:
        {\tt\small lantao@iu.edu} \\
The paper was published in 2017 International Symposium on Robotics Research (ISRR).
}

\maketitle



\vspace{-95pt}

\abstract
{
The solution convergence of Markov Decision Processes (MDPs) can be accelerated by prioritized sweeping of states ranked by their potential impacts to other states.
In this paper, we present new heuristics to speed up the solution convergence of MDPs.  
First, we quantify the level of {\em reachability} of every state using the  Mean First Passage Time (MFPT) and show that such reachability characterization very well assesses the importance of states which is used for effective state prioritization. 
Then, we introduce the notion of {\em backup differentials} as an extension to the prioritized sweeping mechanism, in order to evaluate the impacts of states at an even finer scale.
Finally, we extend the state prioritization to the temporal process, where only partial sweeping can be performed during certain intermediate value iteration stages. 
To validate our design, we have performed numerical evaluations by comparing the proposed new heuristics with corresponding classic baseline mechanisms. The evaluation results showed that our reachability based framework and its differential variants have outperformed the state-of-the-art solutions in terms of both practical runtime and number of iterations. 
}


\vspace{-15pt}
\section{Introduction}
\vspace{-5pt}
Decision-making in uncertain environments is a basic problem in the area of artificial intelligence~\cite{russell02,sigaud2013markov}, and Markov decision processes (MDPs) have become very popular for modeling non-deterministic planning problems with full observability~\cite{puterman2014markov,white1993survey}.
Specifically, an MDP assumes discrete states and discrete actions, and can be viewed as stochastic automata where an agent's actions have uncertain effects. Such uncertain action outcomes induce stochastic transitions between states.
The expected value of a chosen action is a function of the transitions it induces.
On executing the action, the agent receives a reward and also causes a change in the state of the environment. The objective of the agent is to perform actions in order to maximize the cumulative future reward over a period of time. 
In practice, the Value Iteration (VI)  is probably the most famous and most widely used method for solving the MDPs~\cite{howarddynamic60,Bertsekas1987}.

We are interested in the exact solution methods and our objective is to further accelerate the convergence of the MDPs' solving mechanism. 
Different from many popular state-space search based heuristics which essentially exploit MDP's probabilistic transition models and search for a path leading from the start state to the goal state based on some search tree/graph structures, our method evaluates the global feature of the entire state space and quantifies the level of reachability of each states using first passage time information.
This enables us to assess the relevance or importance of states, built on which other completely new heuristics such as {\em backup differentials} can be designed. 

In greater detail, in this  paper we introduce the notion of {\em reachability landscape} which characterizes 
how hard it is for the agent to transit from any state to the given goal state. 
To compute the reachability landscape, we use the Mean First Passage Time (MFPT) which can be formulated into a simple linear system. 
We show that such reachability characterization of each state reflects the importance of this state, and thus provides a natural basis that can be used for prioritizing states for the standard value iteration process.  
With that, we also propose the {\em differential backup} based heuristic where the potential impacts of states can be more accurately captured. 
In addition, we extend the classic prioritized sweeping to the temporal process and re-allocate the sweeping efforts during different stages, so  that only partial sweeping can be performed during certain intermediate value iteration stages and the convergence performance can be further improved. 

\vspace{-20pt}
\section{Related Work} 
\vspace{-5pt}
The basic computational mechanisms and techniques for MDPs have been well-understood and widely applied to solve many decision-theoretic planning~\cite{BoutilierDTP99,sutton1990integrated} and reinforcement learning problems~\cite{busoniu2010reinforcement, van2012reinforcement}. 
Value iteration and policy iteration are two of the most famous algorithms to solve the MDPs and particularly the value iteration might be the most widely used mechanism due to its easy implementation and fast convergence~\cite{howarddynamic60,Bertsekas1987}.

One of the most widely-used frameworks for speeding up the convergence is a rich class of heuristics based on state-space search~\cite{hansen2001lao,Barto1995}. 
Such heuristics usually take advantage of MDPs' probabilistic transition models directly, and use a tree or graph structure to 
search for a solution in the form of a sequence of actions, leading to a path
from the start state to a goal state. 
For instance, the most well known method in this category is probably the {\em real-time dynamic programming (RTDP)}~\cite{Barto1995} where states are not treated uniformly. In each DP iteration of the RTDP, only a subset of most important states might be explored, and the selection of the subset of states are usually built on,  and related to, the agent's exploration experience. 
For a single backup iteration, the RTDP typically requires less computation time in comparison to the classic DP where all states need to be swept, and thus can be extended as an online process and integrated into the real-time reinforcement learning framework~\cite{bonet2003labeled}.
Similar strategies also include the {\em state abstraction}~\cite{andre2002state,li2006towards}, where states with similar characteristics are hierarchically and/or adaptively grouped together, either in offline static or online dynamic aggregation style. 
Different from all these methods, the proposed framework is not based on state-space search, instead it characterizes the entire state space into a reachability landscape using the MFPT information.


Another important heuristic for efficiently solving MDPs is the {\em prioritized sweeping}~\cite{Moore93prioritizedsweeping}, which has been broadly employed to further speed up the value iteration process. 
This heuristic evaluates each state and obtains a score based on the state's contribution to the convergence, and then prioritizes/sorts all states based on their scores (e.g., those states with larger difference in value between two consecutive iterations will get higher scores)~\cite{parr1998generalized,wingate2005prioritization}.  
Then in the immediately next dynamic programming iteration, evaluating the states follows the newly prioritized order. 
The prioritized sweeping heuristic is also leveraged in our MFPT based value iteration procedure, and comparisons with baseline approaches have been conducted in our experimental section.

The reachability of state space has been investigated in existing works. For example, the {\em structured reachability analysis}~\cite{boutilier1998structured} of MDPs has been proposed to evaluate whether a state is reachable or not, so that one can restrict the dynamic programming to only reachable states,
reducing the computational burden of solving an MDP. 
Note, the reachability in that work is defined as a binary state, and if a state is eventually reachable from a given starting state, then it is defined as reachable, otherwise it is unreachable. This is different from our reachability landscape where each state's reachability is measured with a real-valued number.

Important related frameworks for solving MDPs also include compact representations such as linear function representation and approximation~\cite{howarddynamic60,puterman2014markov} used in the policy iteration algorithms. 
The linear equation based techniques do not exploit regions of uniformity in value functions associated with states, but rather a compact form of state features that can somewhat reflect values~\cite{boutilier2000stochastic}. 
Our method for computing the MFPT can also be formulated into a linear system. 
However, the intermediate results generated from MFPT are more direct:  the produced reachability landscape represented by a ``grid map" very well capture -- and also allow us to visualize -- the relevance or importance of states, and can lead to a faster convergence speed which is demonstrated in the experiments.



\definecolor{roweven}{rgb}{1,1,1}
\definecolor{rowodd}{rgb}{0.95,0.95,0.995}

\vspace{-20pt}
\section{Preliminaries}
\vspace{-5pt}
\label{Preliminaries}

 
\vspace{-5pt}
\subsection{Markov Decision Processes}
\vspace{-5pt}

\begin{definition} 
An Markov Decision Process (MDP) is a tuple $M = (S,A,T,R)$, where 
\begin{itemize}
    \item $S = \{s_1, \cdots, s_n\}$ is a set of states;
    \item $A = \{a_1, \cdots, a_n\}$ is a set of actions;
    \item $T : S \times A \times S \rightarrow [0,1]$ is a state transition function  such that $T_{a}(s_1,s_2)$ is the probability that action $a$ in state $s_1$ will lead to state $s_2$;
    \item $R : S \times A \rightarrow \mathbb{R}$ is a reward function where  $R_a(s, s')$ returns the immediate reward received on taking action $a$ in state $s$ that will lead to state $s'$.
\end{itemize}
\end{definition}

If every non-terminal state can eventually enter a terminal state such as a goal/destination state, then such a Markov system is {\em absorbing} by nature ~\cite{boutilier2000stochastic}. In this work, we restrict our attention to absorbing Markov systems so that the agent can arrive and stop at a goal.

 A {\em policy} is of the form $\pi = \{s_1 \rightarrow a_{1}, s_2 \rightarrow a_{2},\cdots, s_{n} \rightarrow a_{n} \}$. We denote $\pi[s]$ as the action associated to state $s$. 
If the policy of a MDP is fixed, then the MDP behaves as a Markov chain~\cite{kemeny1959finite}.

\vspace{-15pt}
\subsubsection{Value Iteration}
\vspace{-5pt}

The Value Iteration (VI) is probably the most widely employed approach to solve MDPs.
It is an iterative procedure that calculates the value (or utility in some literature) of each state based on the values of the neighbouring states until it converges. 
The value $V(s)$ of a state $s$ at each iteration can be calculated by the Bellman equation shown below
\begin{equation} \label{eq:mdp1}
V(s) = \max_{a\in A} \sum_{s'\in S} T_a( s, s') \Big(R_a(s, s')  + \gamma V(s') \Big),
\end{equation}
where $\gamma$ is a reward discounting parameter. The stopping criteria for the algorithm is when the values calculated on two consecutive iterations are close enough, i.e., 
$\max_{s \in S} |V(s) - V'(s)| \leq \epsilon$, 
where $\epsilon$ is an optimization tolerance/threshold value, which determines the level of convergence accuracy.
We call one such iterative update a {\em Bellman backup}.

\vspace{-15pt}
\subsubsection{Prioritized Sweeping}
\vspace{-5pt} 
The standard Bellman backup evaluates values of all states in a sweeping style, following the index of the states stored in the memory. 
To speed up the convergence, a heuristic called {\em prioritized sweeping}  has been proposed and widely used as a benchmark framework for non-domain-specific applications. 
The algorithm labels states as more ``relevant" or more ``important" during a particular iteration, if the change in the state value is higher when compared to its previous iteration. 
The essential idea is that, the larger the value changes, the higher impact that updating that state will change its dependent states, thereby taking a larger step towards convergence. 
Alg.~\ref{algo:ps} shows how the value iteration proceeds in prioritized sweeping (VI-PS). 

\begin{algorithm}
    \caption{Prioritized Sweeping (VI-PS)}
    \label{algo:ps}
    Initialize a priority queue, PQ as empty \\
    $s$ := state with change $\delta$ in $V(s)$   \\     
    Insert $s$ into PQ \\
    \While{PQ is not empty}{
        $s$ := pop highest priority entry from PQ \\
        Update $V(s)$ as per Bellman backup \\
        \ForEach{ predecessor state $s'$ of $s$} {
          Calculate change $\delta'$ in $V(s')$ \\
          \If{$\delta'$ $\geq$ threshold }{
                    priority of $s'$ := max (current priority, $\text{max}_{a}(\delta\cdot T_a(s',s)$) \\
                    Insert or update PQ with $s'$ and the calculated priority \\
                }
        }
    }        
\end{algorithm}


\vspace{-5pt}
\subsection{Mean First Passage Times}
\vspace{-5pt}
Starting from state $s_i$, the number of state transitions involved in reaching states $s_j$ for the first time  is referred as {\em first passage time (FPT)}, $T_{ij}$. The {\em mean first passage time (MFPT)}, $\mu_{ij}$ from state $s_i$ to $s_j$ is the expected number of steps required to transit from state $s_i$ to an absorbing state $s_j$ ~\cite{AssafSharedShanthikumar1985}. 
The MFPT analysis is built on the Markov chain and give
us information about the short range behavior of the chain. Moreover, it has nothing to do with the agent's actions.
Remember that, when an MDP is associated to a policy, it then behaves as a Markov chain~\cite{kemeny1959finite}. 

Formally, let us define a Markov chain with $n$ states and transition probability matrix, $p \in {\rm I\!R}^{n,n}$ where $p_{ik}$ represents the transition probability from state $s_i$ to $s_k$. If the transition probability matrix is regular, then we can refer each MFPT, $\mu_{ij} = E(T_{ij})$. Given, $B_k$ is an event where the first state transition happens from state $s_i$ to $s_k$, MFPT satisfies the below conditional expectation formula:
\begin{equation}\label{eq:fpt00}
E(T_{ij}) = \sum_{k} E(T_{ij} | B_k) p_{ik}
\end{equation}

From the definition of mean first passage times, we have, $E(T_{ij} | B_k) = 1 + E(T_{kj})$. So, we can rewrite Eq.~\eqref{eq:fpt00} as follows.
\begin{equation}\label{eq:fpt01}
E(T_{ij}) = \sum_{k} p_{ik} + \sum_{k \neq j} E(T_{kj}) p_{ik}
\end{equation}
Since, $\sum_{k} p_{ik}$ = 1, Eq.~\eqref{eq:fpt01} can be formulated as per the below equation:
\begin{equation}\label{eq:fpt002}
\mu_{ij} = 1 + \sum_{k \neq j} p_{ik} * \mu_{kj} 
\end{equation}
Eq.~\eqref{eq:fpt002} can be rewritten as:
\begin{equation}
\sum_{k \neq j} p_{ik} * \mu_{kj} - \mu_{ij} = -1,
\end{equation}
Solving all MFPT variables can be viewed as solving a system of linear equations 
\begin{equation}
\label{eq:fpt2}
\begin{spmatrix}{}
    p_{11} - 1 & p_{12} & .. & .. & p_{1n} \\        
    p_{21} & p_{22} - 1 & .. & .. & p_{2n} \\
    .. & .. & .. & .. & .. \\
    .. & .. & .. & .. & .. \\
    p_{n1} & p_{n2} & .. & .. & p_{nn} - 1 \\
\end{spmatrix}
\begin{spmatrix}{}
    \mu_{1j} \\        \mu_{2j} \\ .. \\ .. \\ \mu_{nj}
\end{spmatrix}
=
\begin{spmatrix}{}
    -1  \\        -1 \\ .. \\ .. \\ -1
\end{spmatrix}.
\end{equation}
The values $\mu_{1j}$, $\mu_{2j}$, $....$, $\mu_{nj}$ represent the MFPTs calculated for state transitions from states $s_1$, $s_2$, $....$, $s_n$ to $s_j$. 
To solve above equation, efficient decomposition methods~\cite{Golub1996} may help to avoid a direct matrix inversion.

\vspace{-15pt}
\section{Reachability and Differential based Solution}
\label{technical}
\vspace{-5pt}

Our objective is to accelerate the convergence process of solving an MDP and we propose a reachability and differential based solution.
Specifically,  we first quantify the level of {\em reachability} of every state using the Mean First Passage Time (MFPT); 
then, we introduce the notion of {\em backup differential} as an extension to capture the state's potential impact to other states at an even finer level. 
Finally, the reachability and differential based mechanism is integrated with the temporally prioritized partial-space sweeping heuristics.

\begin{figure}[t]
  \centering
  \subfigure[]
        {\label{fig:SI_Env}\includegraphics[height=1.3in]{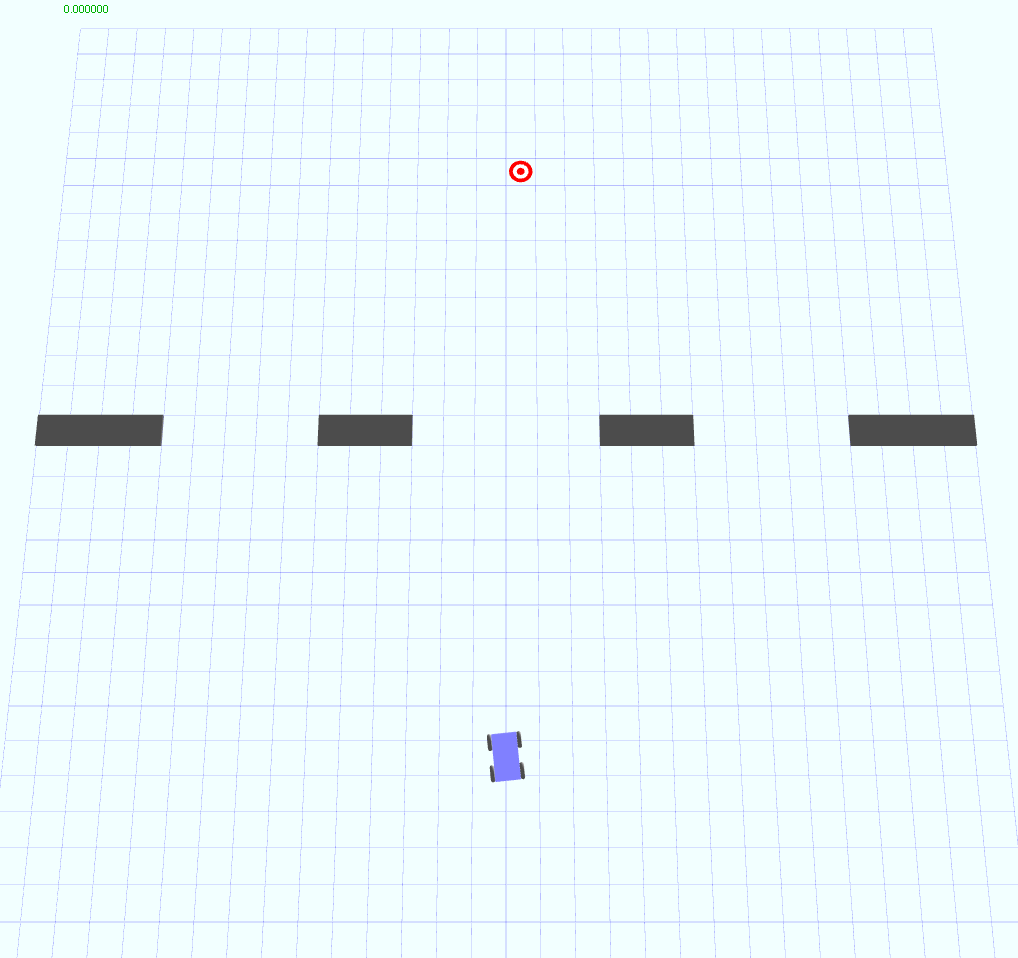}} \quad \quad
  \subfigure[]
        {\label{fig:heatmap3}\includegraphics[height=1.3in]{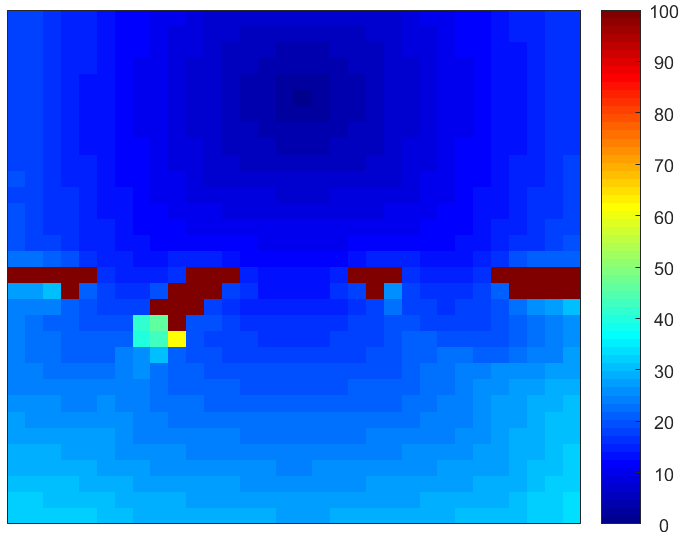}}
	\vspace{-10pt}
	\caption{\small Illustration of reachability landscape. (a) Demonstration of a simple simulation scenario with dark blocks as obstacles, and the goal state as a red circle; (b) Final reachability landscape corresponding to the converged MDP.}
	\label{fig:heatmap}
	\vspace{-5pt}
\end{figure}

\vspace{-15pt}
\subsection{Construction of Reachability Landscape}
\vspace{-5pt}

By ``reachability of a state" we mean that based on current fixed policy, how hard it is for the agent to transit from this state to the given goal/absorbing state. 
A smaller reachability value is defined as more reachable.
Thus, a state's reachability in this context is in fact a real-valued number instead of a binary value (reachable vs. unreachable), and for all states we can construct a {\em reachability landscape}, which can be visualized as a grid map if the states have some spatial structure, see Fig.~\ref{fig:heatmap} for an illustration.

Specifically, assume the goal state is $s_j$, we use the MFPT value $\mu_{ij}$ to quantify the reachability from an arbitrary state $s_i$ to the goal state $s_j$.
For all states, we can then construct a reachability landscape represented as a surface.
Fig.~\ref{fig:heatmap3} shows an example represented as heatmap in our simulated environment. 
The values in the heatmap range from 0 (cold color) to 100 (warm color). 
Intuitively, one may imagine the landscape to be an ``energy surface" where the minimum energy state is the most stabilized, and an agent situating at a non-minimum state will be unstable and eventually moves towards to the minimum state, which is the absorbing state where the goal is located.
It is worth mentioning that, in order to better analyze the low MFPT spectrum that we are most interested, values in reachability landscape may be ``clipped". 
For example, in Fig.~\ref{fig:heatmap3} any value greater than 100 has been clipped to 100. 

We observe that the reachability  conveys very useful information on potential impacts to other states. 
More specifically, a state with a better reachability (smaller MFPT value) is more likely to make a larger change during the MDP convergence procedure, leading to a bigger convergence step.

\vspace{-15pt}
\subsection{Mean First Passage Time based Value Iteration (MFPT-VI)}
\vspace{-5pt}

The {\em prioritized sweeping} mechanism has further improved the convergence rate over the classic Bellman backup by exploiting and ranking the states based on their potential contributions, where the metric is based on comparing the difference of values for each state between two consecutive iterations.

We propose a new method called Mean First Passage Time based Value Iteration (MFPT-VI) which is also built on the prioritized sweeping mechanism, but using a different prioritizing metric. 
Formally, we propose to prioritize the states using the reachability values, because as aforementioned, the reachability characterization of each state reflects the potential impact/contribution of this state, and thus provides a natural basis for prioritization. 
Our reachability based metric is distinct from the existing value-difference metric as follows,
the reachability landscape can very well capture the degree of importance for all states from a global viewpoint, whereas the classic value-difference strategy evaluates states locally, and may fail to grasp the correct global ``convergence direction" due to the local viewpoint.

Note that, since the MFPT computation is relatively expensive and it is particularly good at capturing high-level feature, it is not necessary to compute the MFPT at every iteration, but rather after every few iterations.
It is also worth mentioning that, if there are multiple goal/absorbing states, then each goal state will require to compute its own MFPT landscape, and the final reachability landscape will be normalized across all obtained landscapes. 
The computational process of MFPT-VI is pseudo-coded in Alg.~\ref{algo:fptvi}.

\vspace{-10pt}
\begin{algorithm}
    \caption{Mean First Passage Time based Value Iteration (MFPT-VI)}
    \label{algo:fptvi}
    {\small
        Given states $S$, actions $A$, transition probability $T_{a}(s,s')$ 
        and reward $R_{a}(s,s')$. Assume goal state $s^*$,
        calculate the optimal policy $\pi$\\
                
        \While{true}{
            $V = V'$ \\
            
            
            Calculate MFPT values $\mu_{1s^*}$, $\mu_{2s^*}$, $\cdots$, $\mu_{|S|s^*}$ by solving the linear system as shown in Eq. ~\eqref{eq:fpt2} \\
            
            List $L$ := Sorted states with increasing order of MFPT values \\
            
            \ForEach{state $s$ in $L$} {
                Compute value update at each state $s$ given policy $\pi$:  
                $V'(s) = \max_{a\in A} \sum_{\forall{s'\in S}} T_a( s, s') \Big(R_a(s, s')  + \gamma V(s') \Big)$
            }
                        
            \If{$\max_{s_i} |V(s_i) - V'(s_i)| \leq \epsilon$}{
                break \\ 
            }
          }
        }
\end{algorithm}
\vspace{-5pt}


\vspace{-10pt}
\subsection{A Heuristic using Backup Differentials}
\vspace{-5pt}


Since the reachability of a state implies how hard it is to reach the given goal state from that state, and during the value iteration process the reachability landscape is re-computed periodically representing a more refined overview of the reachability characterization, therefore the difference from the old landscape to an updated version can indicate the ``changing direction of the reachability" to some extent. Intuitively, a larger error (reduction of the reachability value) on a state implies more significant potential that this state can impact the convergence.  
Since the error is the difference between two subsequent landscapes and it occurs during the Bellman backup, we call such error computation a  {\em backup differential}. 

We propose to extend the MFPT-VI to use the backup differential as a prioritizing metric, and term this method D-MFPT-VI, where ``D" means differential.
Based on above observation, the potential impact of a state can also be better captured by the {\em rate of error changes}, which is essentially the higher order of differentials.
For example, the next order of the backup differential is D2-MFPT-VI, accounting for the possible future trends of the error, based on its current rate of change.
This is analogous to the kinematics analysis of displacement, velocity and acceleration, etc.

Obviously, the standard VI-PS, that only  measures the error between two iterations is a special case of such differential framework. 
In addition to that,  we also implemented a one-order higher differential version of VI-PS and we refer it to as D-VI-PS, and compared its performance with the  reachability based variant of the equivalent order of backup differential.
The computational process of D-MFPT-VI is pseudo-coded in Alg.~\ref{algo:dfptvi}.




\vspace{-5pt}
\begin{algorithm}
    \caption{Differential Mean First Passage Time based Value Iteration (D-MFPT-VI)}
    \label{algo:dfptvi}
    {\small
        Given states $S$, actions $A$, transition probability $T_{a}(s,s')$ 
        and reward $R_{a}(s,s')$. Assume goal state $s^*$,
        calculate the optimal policy $\pi$\\
                
        \While{true}{
            $V = V'$ \\
            
            
            $MFPT_{Old}$ = Current MFPT values $\mu_{1s^*}$, $\mu_{2s^*}$, $\cdots$, $\mu_{|S|s^*}$ \\
            Calculate MFPT values $\mu_{1s^*}$, $\mu_{2s^*}$, $\cdots$, $\mu_{|S|s^*}$ by solving the linear system as shown in Eq. ~\eqref{eq:fpt2} \\
            
            $MFPT_{New}$ = Newly calculated MFPT values $\mu_{1s^*}$, $\mu_{2s^*}$, $\cdots$, $\mu_{|S|s^*}$ \\
            
            $\Delta_{MFPT}$ = $MFPT_{New}$ - $MFPT_{Old}$ \\
            
            List $L$ := Sorted states with decreasing order of $\Delta_{MFPT}$ values \\
            
            \ForEach{state $s$ in $L$} {
                Compute value update at each state $s$ given policy $\pi$:  
                $V'(s) = \max_{a\in A} \sum_{\forall{s'\in S}} T_a( s, s') \Big(R_a(s, s')  + \gamma V(s') \Big)$
            }
                        
            \If{$\max_{s_i} |V(s_i) - V'(s_i)| \leq \epsilon$}{
                break \\ 
            }
          }
        }
\end{algorithm}
\vspace{-5pt}


\subsection{Solving MDPs in Partial State Space}
\label{sect:partial}
\vspace{-5pt}

Until here, what we discussed are for prioritized sweeping in full state space, and our experimental results in Section~\ref{Results} will show that the proposed framework is superior to the baseline popular methods. 
In this section, we show that by extending the framework to partial space sweeping, the convergence rate can be further enhanced.

Our heuristics are based on the allocation of sweeping efforts.
Specifically, the prioritized sweeping mechanism has re-allocated the state sweeping from an arbitrary set to a sorted list, but such re-allocation is still limited to one iteration at a particular time moment.
Because the solving process involves many iterations (and moments), we take advantage of the temporal process and re-allocate the sweeping efforts during different temporal stages.
In greater detail, as aforementioned, the reachability characterization at earlier stages can only capture a big picture of the problem and it begins to refine local details in later phases, thus, 
at earlier value iterations we opt to sweep a small subset of states of most impact, and gradually enlarge the sweeping set by including more states of smaller impact, until all states are included in the later stages. 
One challenge here is how to divide the space from the prioritized list.
We provide two heuristics as follows:



\begin{enumerate}[1)]
    
\item The first heuristic is to uniformly divide the prioritized states into sub-lists of equal size (length). If the number of partitions is $p$, and the sweeping method is D-MFPT-VI (for example), then each partitioned sub-list contains $|S|/p$ states. We refer to this heuristic as D-MFPT-VI-H1.

\item The second heuristic is to uniformly divide the ``impact", but in this way the lengths of sub-lists can be non-uniform. 
Formally, given the number of partition $p$ and the errors obtained in D-MFPT-VI, and assume the maximum error (corresponding to largest impact) and minimum error (corresponding to smallest impact) are  $e_{max}$ and $e_{min}$, respectively. Then the error range can be calculated as $r = e_{max}$ - $e_{min}$ and we wish each partitioned sub-list to have an error range of $\frac{r}{p}$.
Consequently, for partitioning index $i= 1, \cdots, p$, a state with an error between $e_{min} + (i-1)*r/p$ and  $e_{min} + i*r/p$ belongs to partition $i$. We refer to this heuristic as D-MFPT-VI-H2.

\end{enumerate}


\vspace{-15pt}
\subsection{Time Complexity Analysis}
\vspace{-5pt}
Each value iteration has a time complexity of $O(|A||S|^2)$ where $|S|$ denotes the number of states and $|A|$ represents the number of actions. 
Calculation of the MFPT needs to solve a linear system that involves matrix inversion (the matrix decomposition has a time complexity of $O(|S|^{2.3})$ if state-of-the-art algorithms are employed~\cite{Golub1996}, given that the size of matrix is the number of states $|S|$). 
Therefore, for each iteration, both the MFPT-VI and D-MFPT-VI algorithms have a time complexity $O(|A||S|^2 + |S|^{2.3})$.
However, experimental results show that our reachability characterization based framework practically requires much fewer iterations to converge. 



\vspace{-5pt}
\section{Experimental Results}
\vspace{-5pt}
\label{Results}


We validated our method through numerical evaluations with a simulator written in C++ running on a Linux machine.

\begin{figure}[t]
  \centering
  \subfigure[]
        {\label{fig:S_Env}\includegraphics[height=1.1in]{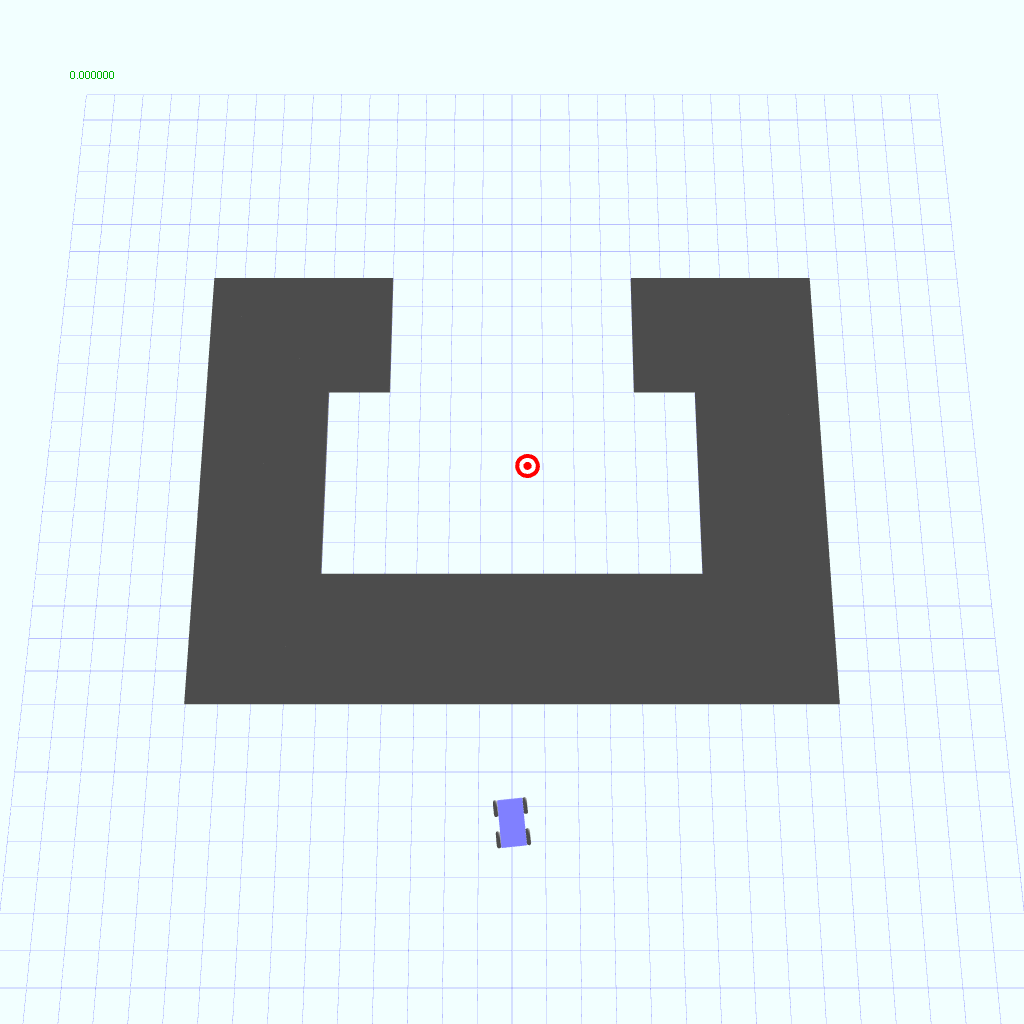}}
  \subfigure[]
        {\label{fig:heatmap01}\includegraphics[height=1.1in]{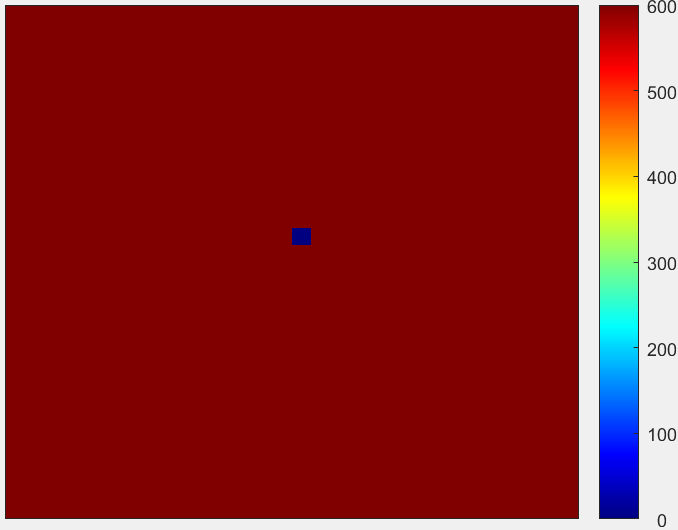}}
  \subfigure[]    
        {\label{fig:heatmap11}\includegraphics[height=1.1in]{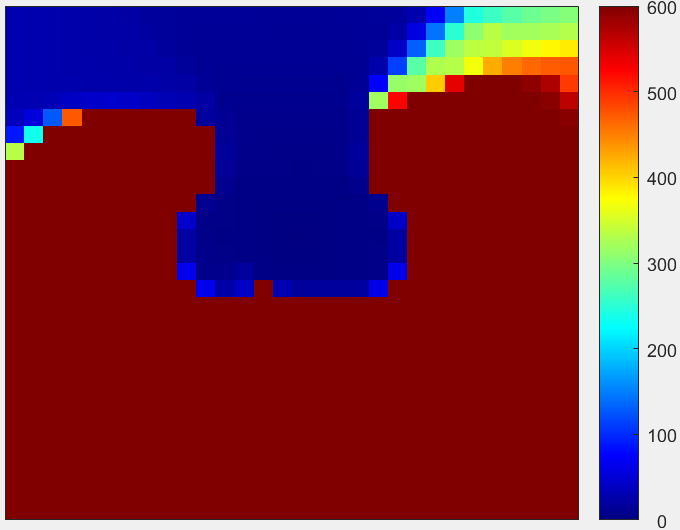}} 
  \subfigure[]
        {\label{fig:heatmap21}\includegraphics[height=1.1in]{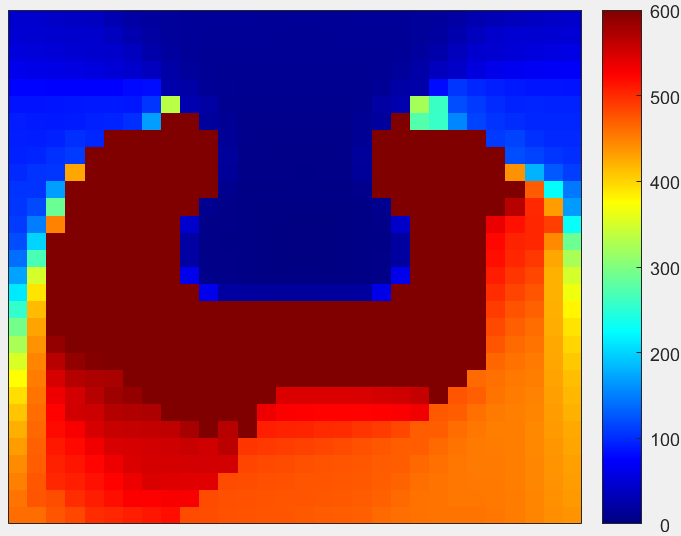}}     
  \subfigure[]
        {\label{fig:heatmap41}\includegraphics[height=1.1in]{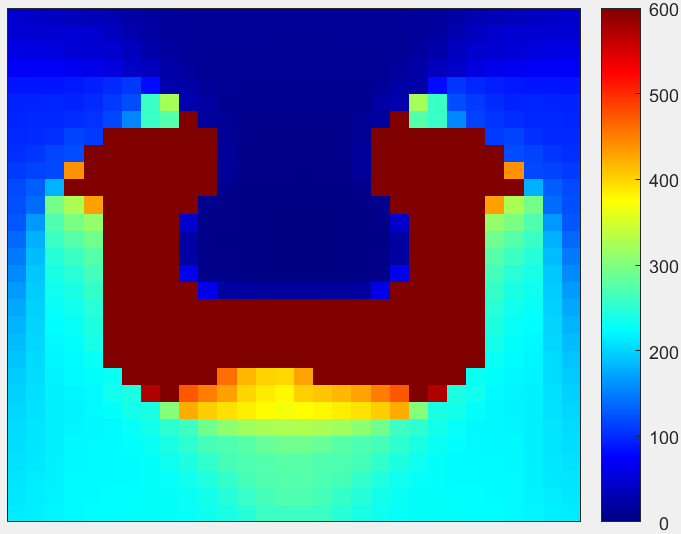}}
  \subfigure[]
        {\label{fig:trajectory}\includegraphics[height=1.1in]{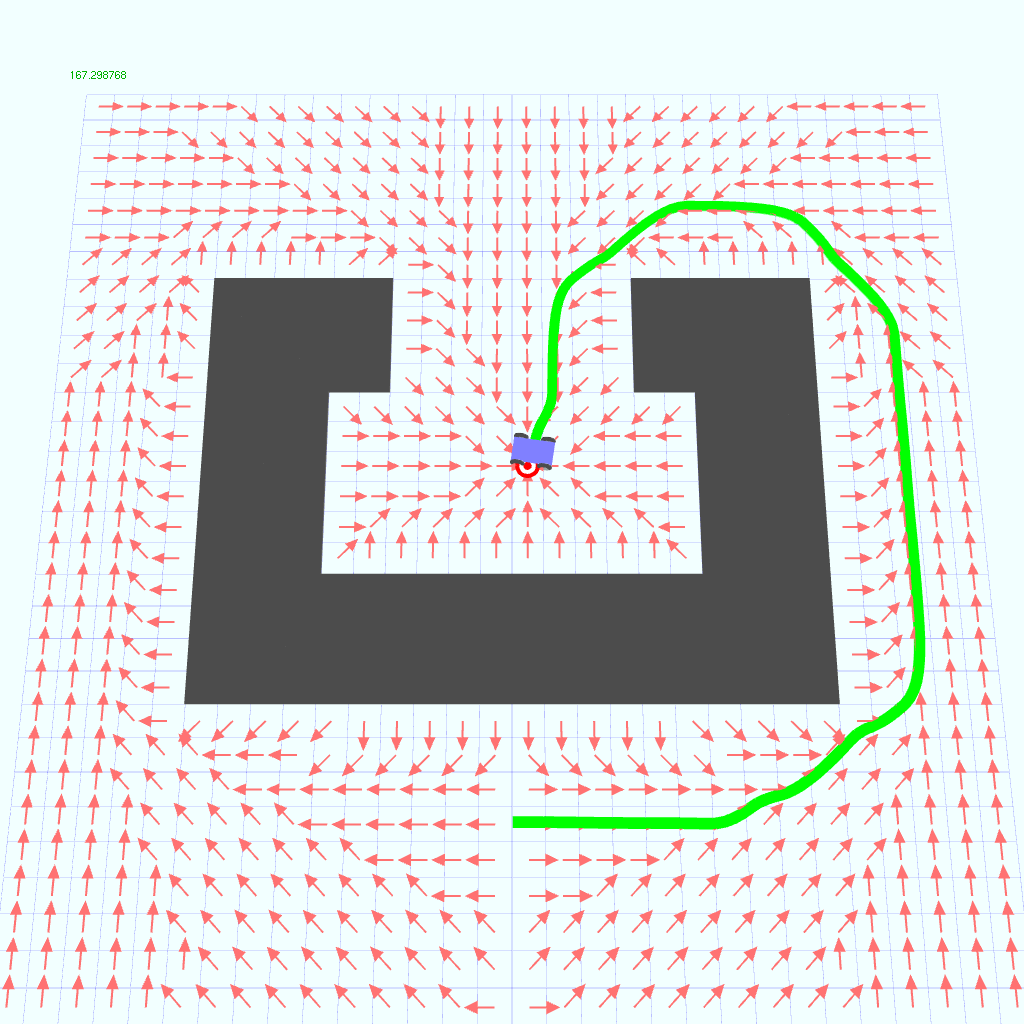}}
	\vspace{-7pt}
	\caption{\small (a) Demonstration of simulation environment, with the agent's initial state (blue) and the goal state (red). Grey blocks are obstacles; (b)-(e) Evolution of  reachability landscapes; (f) Converged optimal policy (red arrows) and a trajectory completed by the agent to reach the goal.}
	\label{fig:heatmapApp}
	\vspace{-10pt}
\end{figure}

We consider the generic MDP problem where each action can lead to transitions to all other states with certain transition probabilities.
However, in many practical scenarios, the probability of transiting from a state to another state that is ``weakly connected" can be small, even close to 0. This can potentially result in non-dense transition matrix.
For example, in the robotic motion planning scenario, the state transition probability from state $s_i$ to state $s_j$ can be correlated with the time or distance of traveling from $s_i$ to $s_j$, and it is more likely for a state to transit to some states within certain vicinity.  
To obtain the discrete MDP states, we tessellate the agent's workspace into a grid map, and represent the center of each grid as a state.
In this way, the state hopping between two states represents the corresponding motion in the workspace.
A demonstration is shown in Fig.~\ref{fig:heatmapApp}.
All experiments were performed on a laptop computer with 8GB RAM and 2.6 GHz quad-core Intel i7 processor.

\begin{figure}[t]
  \centering
  \subfigure[]
        {\label{fig:TS_V}\includegraphics[height=1.3in]{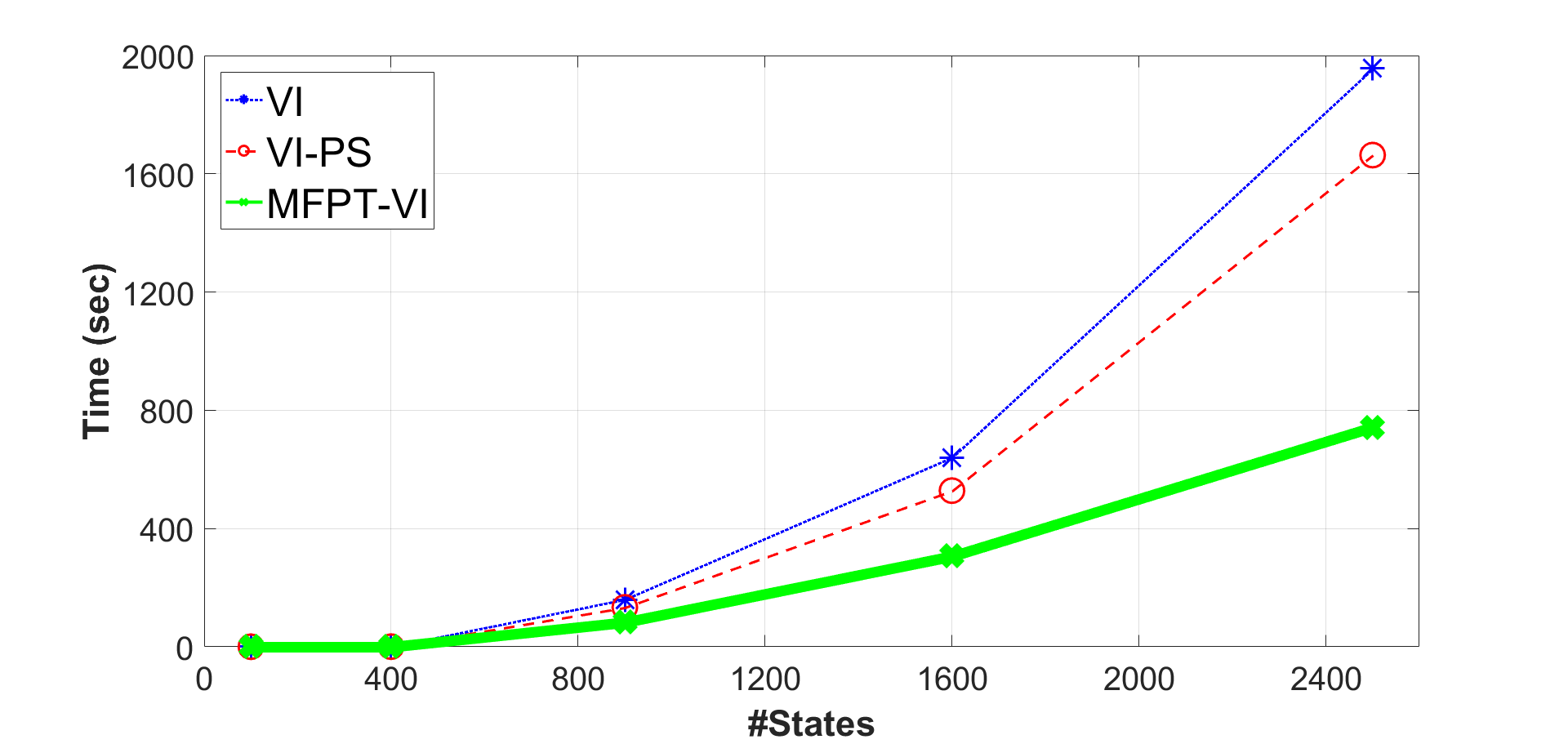}}
        \quad
  \subfigure[]    
        {\label{fig:TS_P}\includegraphics[height=1.3in]{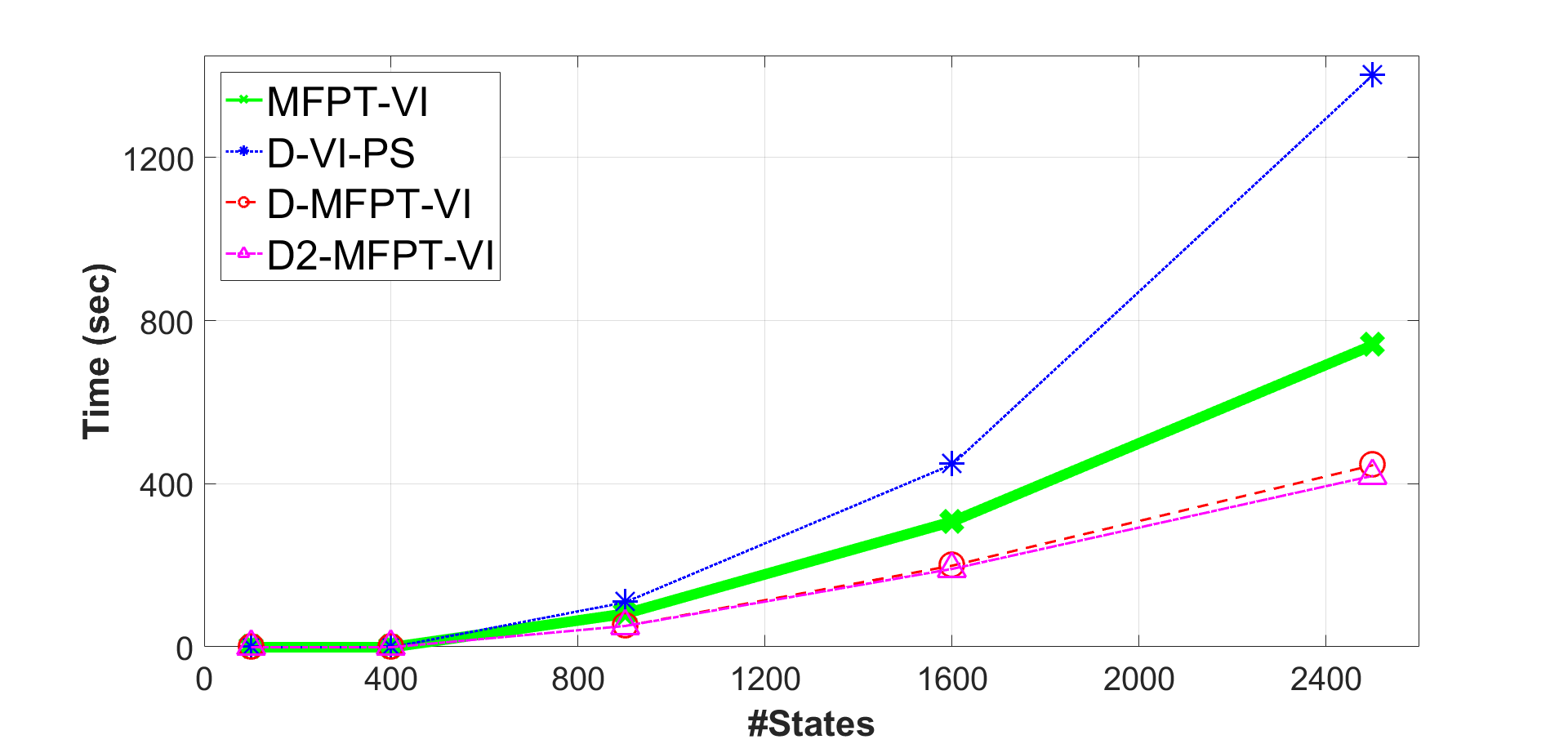}}  
	\vspace{-10pt}
	\caption{\small Time comparisons between the baseline methods and our proposed algorithms, with changing numbers of states ($x$-axis). (a) Variants of value iteration methods. (b) Backup differential variants. The thick green curve in two figures is the result of MFPT-VI, which can be used as a common baseline to compare.
	}
	\label{fig:TimeVsStates}
	\vspace{-15pt}
\end{figure}

\vspace{-15pt}
\subsection{Solution with Full Space Sweeping}
\vspace{-5pt}
To assess the advantages introduced by the MFPT reachability characterization, we first investigate the solution convergence using this method in the full state space, by comparing with other popular full space sweeping mechanisms.  

We start by comparing the practical runtime performance as it is the most basic algorithmic evaluation metric. 
The time taken by VI, VI-PS and MFPT-VI algorithms to converge to the optimal solution (with the same convergence error threshold) are shown in Fig.~\ref{fig:TS_V}. 
The results reveal that VI is the slowest amongst all the three algorithms. 
Because of the prioritized sweeping heuristic, VI-PS is faster than VI. 
It is obvious to see that our proposed MFPT-VI is much faster than the other two methods. 
It is worth noting that in our implementation of MFPT-VI, the MFPT component is computed every three iterations instead of every single iteration, because the the purpose of using MFPT is to characterize global reachability feature which requires less frequent computations. 

Fig.~\ref{fig:TS_P} compares the time taken by MFPT-VI, D-VI-PS, D-MFPT-VI, and D2-MFPT-PI. Firstly, D-VI-PS is faster than the state-of-the-art VI-PS. Moreover, our proposed algorithms based on the reachability abstraction are faster than D-VI-PS.
The differential versions of MFPT-VI i.e. D-MFPT-VI and D2-MFPT-VI are faster compared to MFPT-VI. This clearly establishes the notion of backup differential as an effective mechanism towards runtime improvement.
The results also show that the D2-MFPT-VI is the fastest compared to the other algorithms. Although the improvement in performance is significant on comparing D-MFPT-VI and MFPT-VI, but its not that significant on comparing D-MFPT-VI and D2-MFPT-VI. 
We believe that the differential of second order is sufficient in capturing a state's potential impact, and therefore, we do not analyze beyond the differential of second order.

\begin{figure}[t]
  \centering
  \subfigure[]
        {\label{fig:CS_V}\includegraphics[height=1.3in]{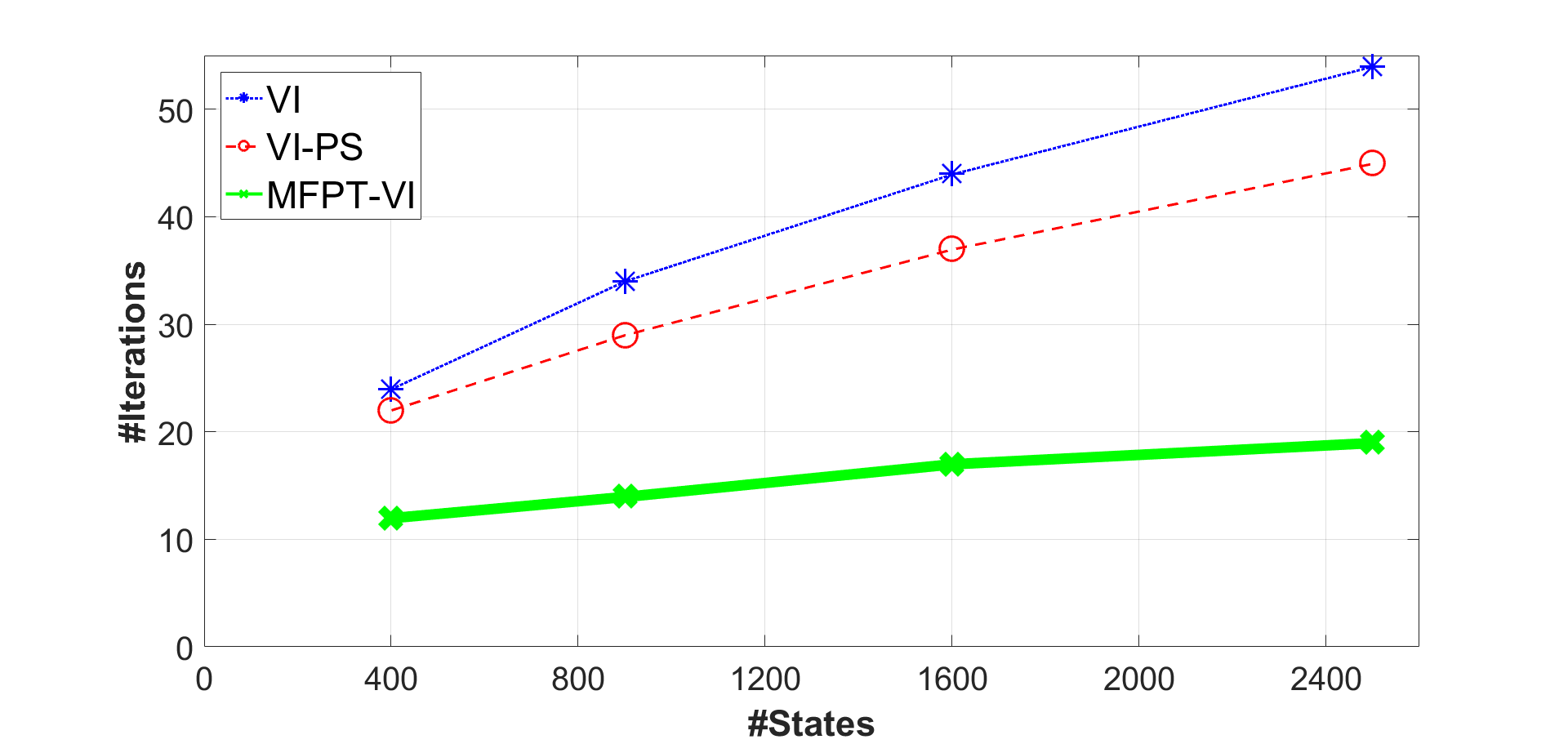}}
        \quad
  \subfigure[]    
        {\label{fig:CS_DV}\includegraphics[height=1.3in]{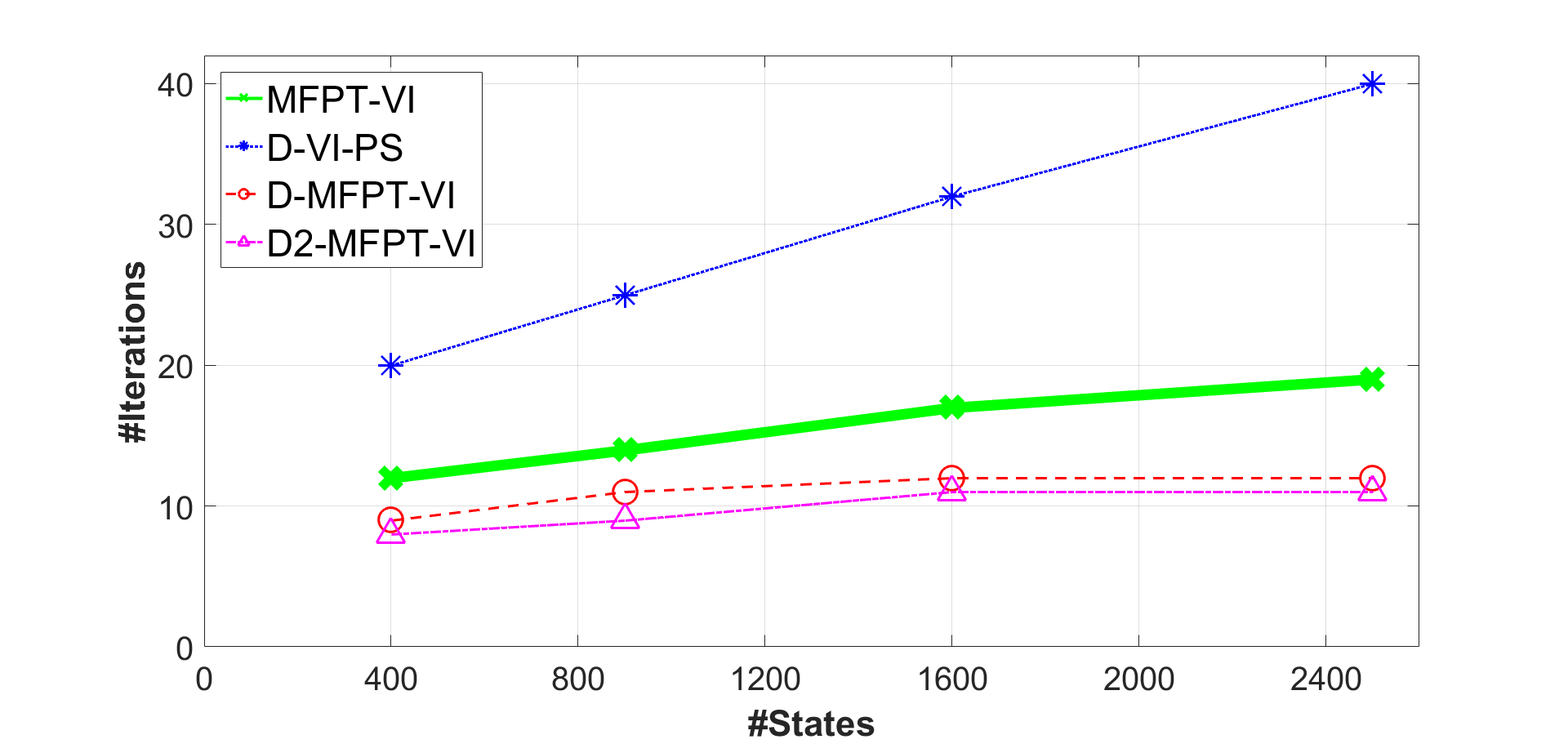}}	\vspace{-10pt}
	\caption{\small Number of iterations required to converge. 
	The $x$-axis denotes the numbers of states. 
	(a) Variants of value iteration methods. (b) Backup differential variants.
	The thick green curve in both figures is the result of MFPT-VI.
	}
	\label{fig:ConvergenceVsStates}
	\vspace{-5pt}
\end{figure}

Next, we evaluate the number of iterations taken by the algorithms to converge to the optimal policy as the number of states changes.
Fig.~\ref{fig:CS_V} compares the number of iterations taken by VI, VI-PS and MFPT-VI, respectively. 
Again, the advantage of our algorithm is obvious, and the results show that the MFPT-VI requires much smaller number of iterations compared to the other two algorithms. 

Fig.~\ref{fig:CS_DV} compares the number of iterations taken by MFPT-VI, D-VI-PS, D-MFPT-VI, and D2-MFPT-VI, from which we can observe that the D2-MFPT-PI converges the fastest amongst all four algorithms. Similar to the runtime behavior shown above, the differential variants -- D-MFPT-VI and D-VI-PS -- converge faster than the non-differential MFPT-VI and VI-PS.
This also implies the remarkable merit of backup differentials as means for faster convergence.


\begin{figure}[t]
  \centering
  \subfigure[]
        {\label{fig:P_VI}\includegraphics[height=1.3in]{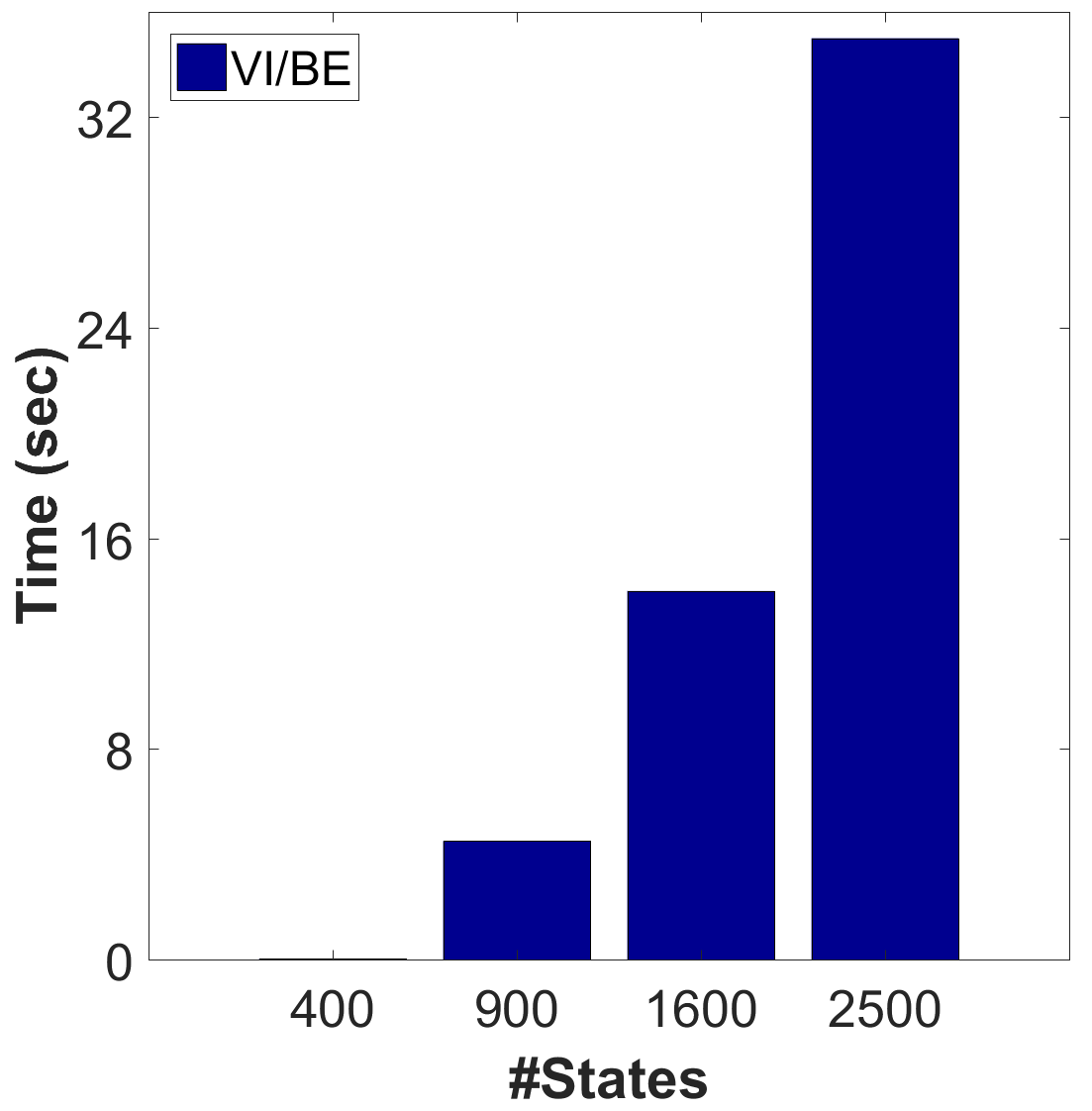}}
  \subfigure[]
        {\label{fig:P_VI_PS}\includegraphics[height=1.3in]{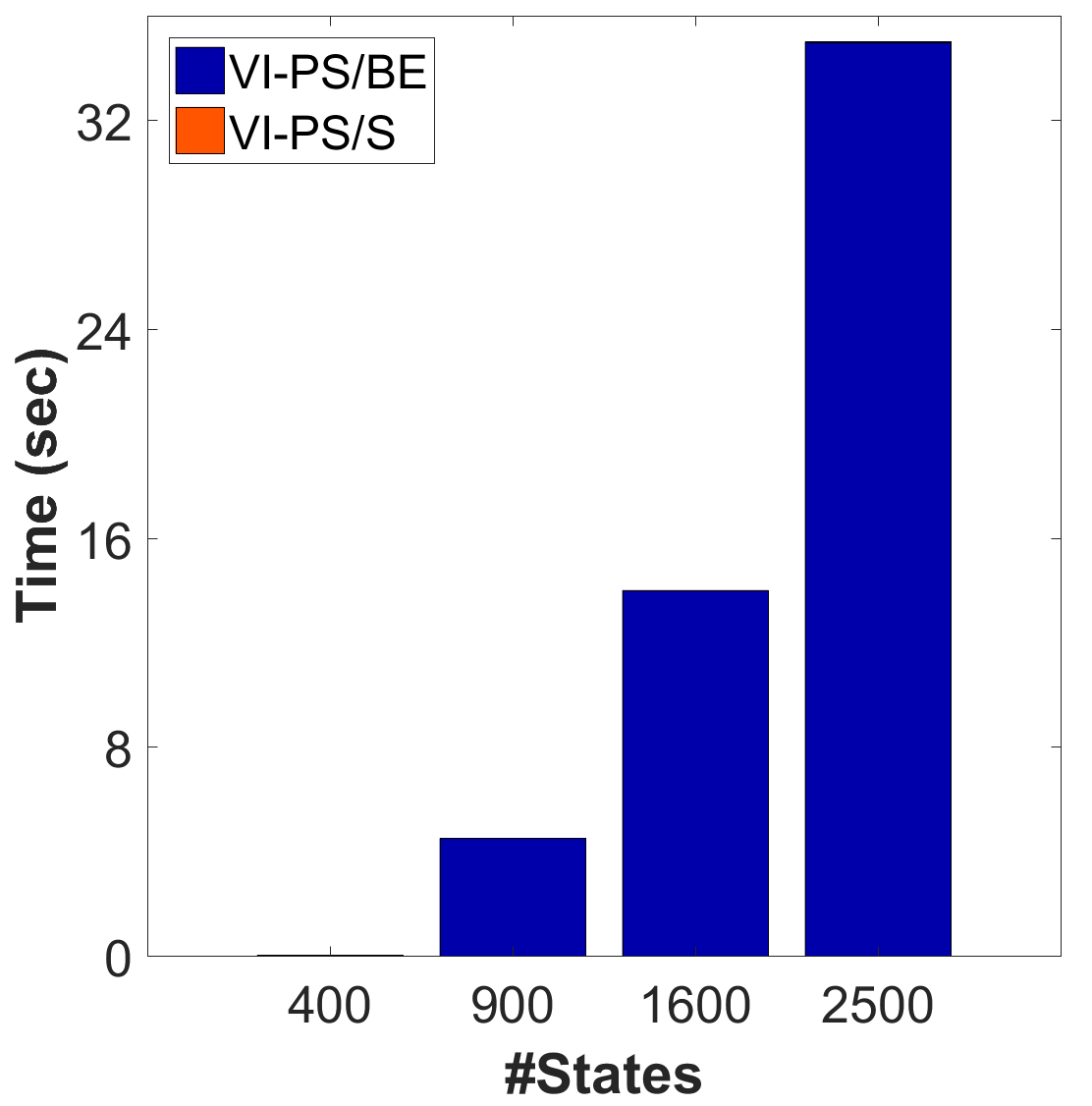}}
  \subfigure[]
        {\label{fig:P_VI_FPT}\includegraphics[height=1.3in]{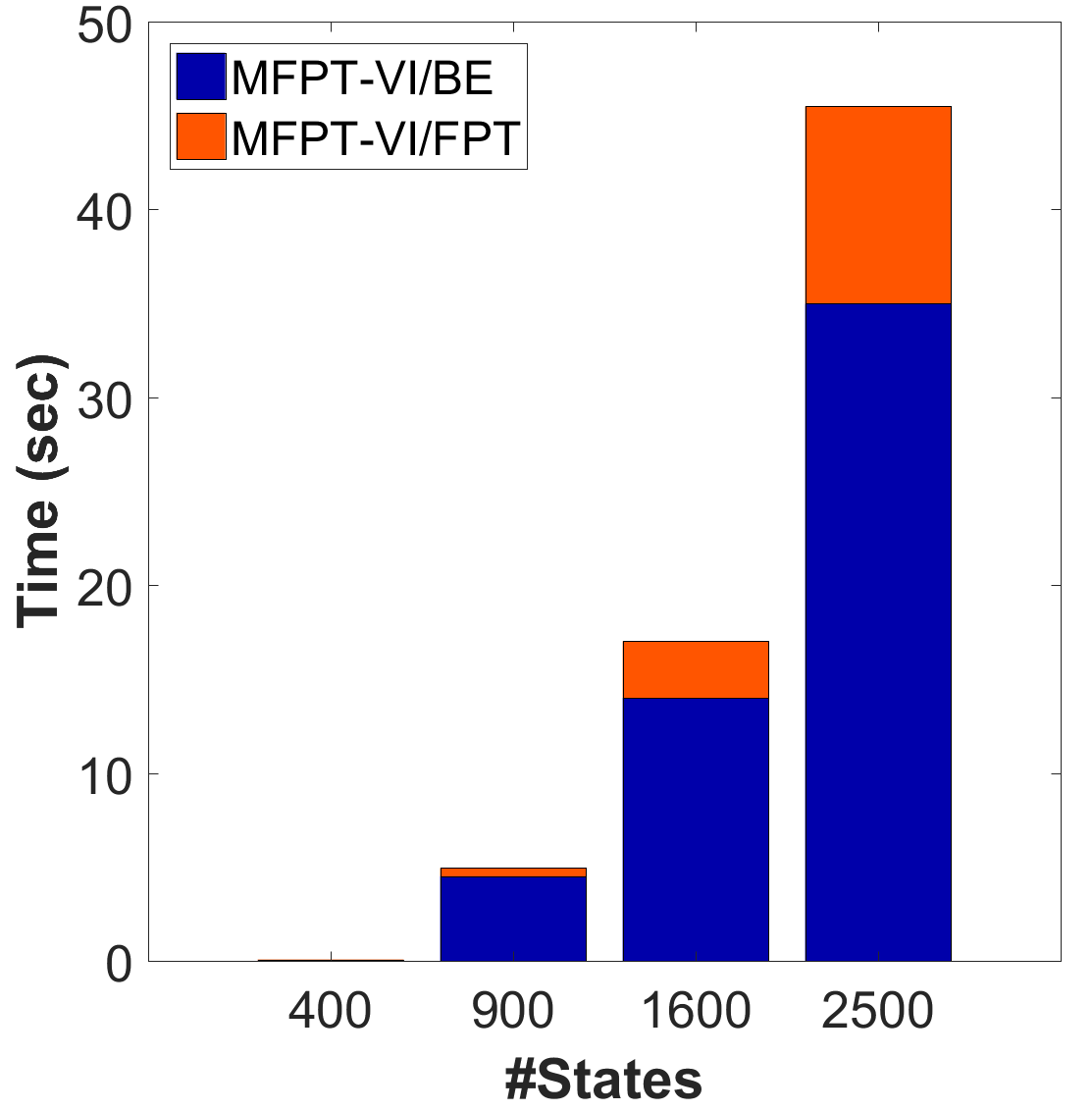}}
	\vspace{-5pt}
	\caption{\small Time taken by individual components of algorithms: (a) VI (b) VI-PS (c) MFPT-VI. In the figures, VI/BE, VI-PS/BE and MFPT-VI/BE represent the Bellman Equation component of VI, VI-PS and MFPT-VI respectively. VI-PS/S denotes the sorting component of VI-PS and MFPT-VI/FPT represents the component that computes MFPT values in the MFPT-VI algorithm. }
	\vspace{-10pt}
\end{figure}


Then, we look into the detailed time taken by the critical components of each algorithm.
A good understanding for the cost of individual component will allow us to better design state sweeping heuristics.
It can be observed from Fig.~\ref{fig:P_VI} that most time is used for Bellman backup operations in the case of VI;
there is an additional step involved for sorting the states for the purpose of states prioritization in case of VI-PS, which is negligible in comparison to the cost of Bellman backup, as presented in Fig.~\ref{fig:P_VI_PS}.
In contrast, majority of time is used for computing the MFPT values in our MFPT-VI algorithm,  as shown in Fig.~\ref{fig:P_VI_FPT}.
Consequently, for the MFPT-VI related algorithms we suggest to update the MFPT result every few iterations instead of re-computing it every iteration.

\begin{figure}[t]
  \centering
  \subfigure[]
        {\label{fig:C_V}\includegraphics[height=1.1in]{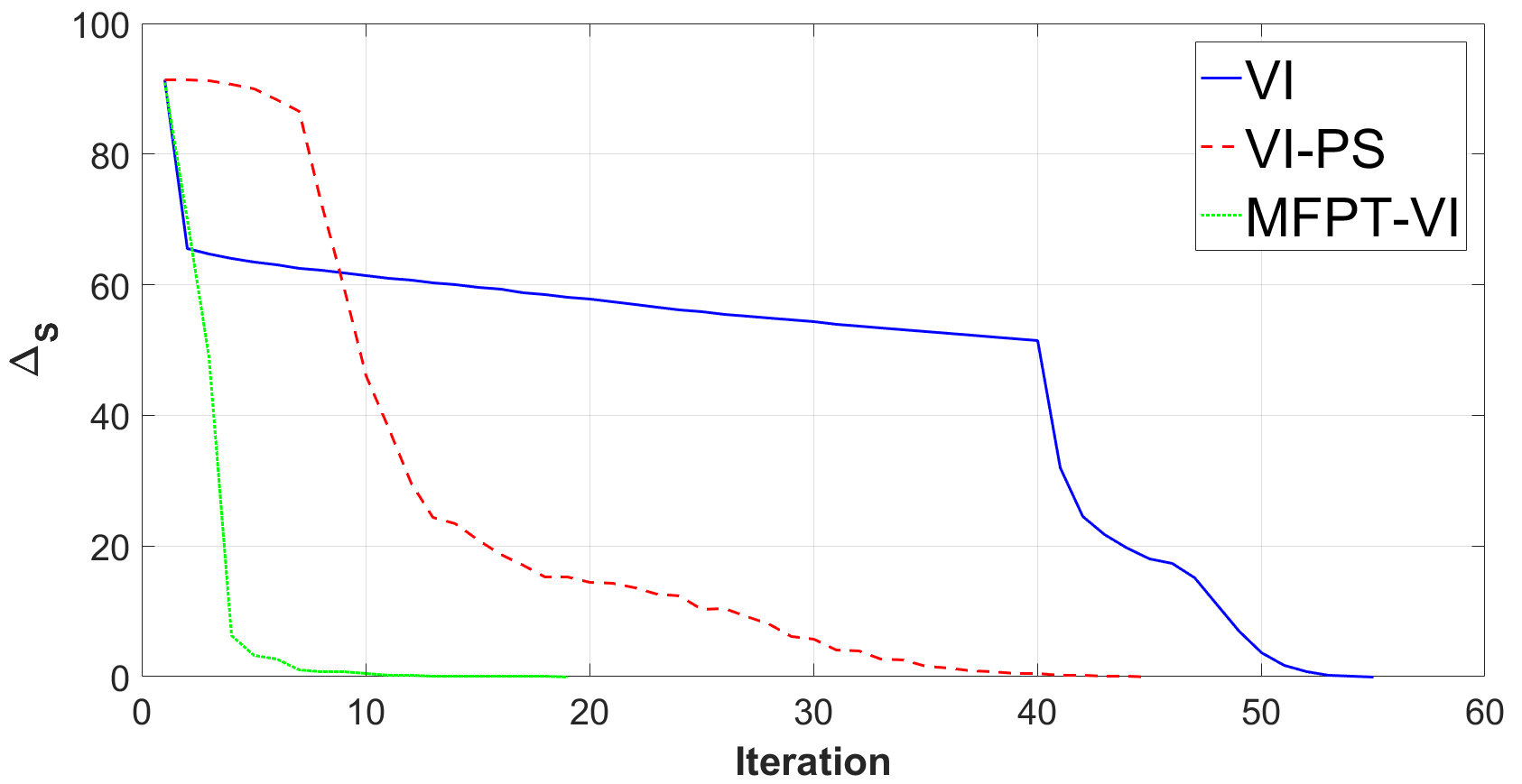}}
        \quad
  \subfigure[]    
        {\label{fig:C_DV}\includegraphics[height=1.1in]{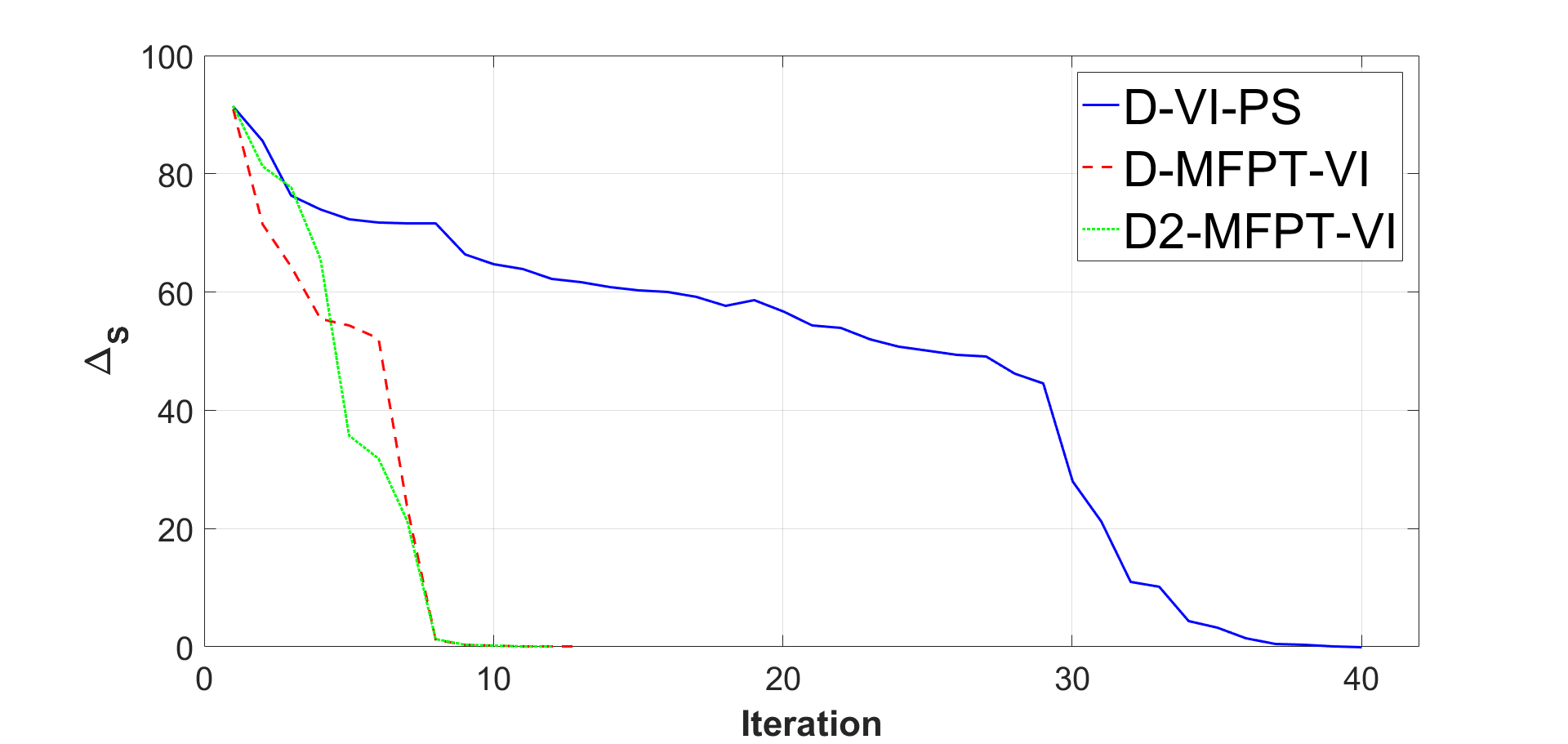}}
	\vspace{-5pt}
	\caption{\small (a) The progress of VI, VI-PS and MFPT-VI across iterations. (b) The progress of D-VI-PS, D-MFPT-VI and D2-MFPT-VI across iterations.}
	\label{fig:progress}
	\vspace{-15pt}
\end{figure}


Finally, we are interested in the convergence curve of each algorithm, through which we can learn and compare the detailed convergence behavior. 
An important criterion to judge the convergence is to see if the difference between two consecutive iterations is small enough. 
Thus, we utilize the maximal error $\Delta_S$ across all states as an evaluation metric. 
Specifically, 
$\Delta_S = \max_{s_i \in S} |V(s_i) - V'(s_i)|$
where $V$, $V'$ represent the values of states at iteration $i$ and $i+1$.

Fig.~\ref{fig:C_V} shows that initially VI, VI-PS and MFPT-VI start with the same $\Delta_S$ value. 
In the first few iterations, our MFPT-VI approach achieves an extremely steep descent compared to the non-reachability based VI and VI-PS. 
For example, in our simulation scenario (with around 2500 states and a threshold value of 0.1), 
the MFPT-VI converges using only 19 iterations; in contrast, VI takes 55 iterations and VI-PS takes 45.

Fig.~\ref{fig:C_DV} shows profiles for D-VI-PS, D-MFPT-VI and D2-MFPT-PI. 
Initially, all the three algorithms start with the same $\Delta_S$ value. 
However, both D2-MFPT-VI and D-MFPT-VI have extremely steep descents in the first few iterations as compared D-VI-PS.
For instance, in the same simulation example mentioned above, our D2-MFPT-VI method converges in 12 iterations whereas D-MFPT-VI takes 13 iterations; in contrast, D-VI-PS takes 40 iterations. 

\begin{figure}[t]
  \centering
  \subfigure[]
        {\label{fig:plotsub0}\includegraphics[height=1.3in]{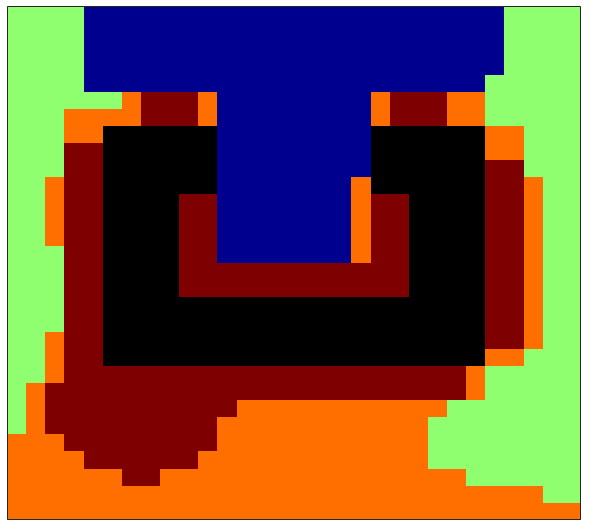}}
        \quad \quad
    \subfigure[]
        {\label{fig:plotsub1}\includegraphics[height=1.3in]{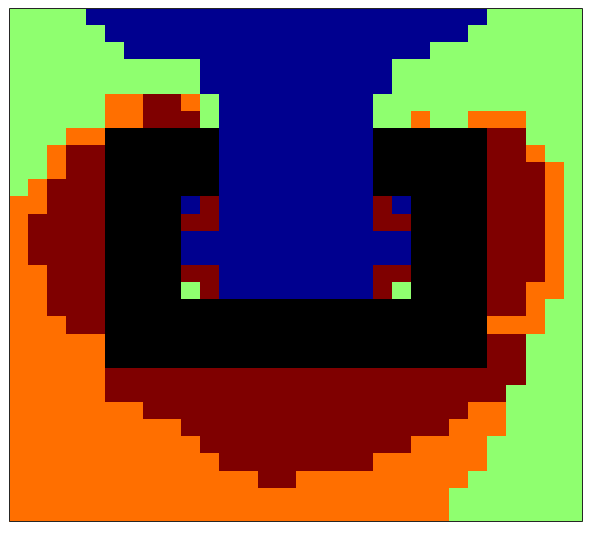}}
	\vspace{-10pt}
	\caption{\small Illustration of partitioned sub-spaces using the example of Fig.~\ref{fig:heatmapApp}. There are four  partitions with different colors and a colder color represents better reachability. The black area in each figure denotes obstacle. (a) Captured partitions at an earlier iteration; (b) Captured partitions at a later iteration.
	}
	\label{fig:plotsubstates}
	\vspace{-5pt}
\end{figure}

\vspace{-15pt}
\subsection{Solution with Partial Space Sweeping}
\vspace{-10pt}
As discussed in Section~\ref{sect:partial}, exploring state space and sweeping in partial space can further accelerate the convergence.
Since the results in full state space reveal that the reachability based algorithms are faster than other algorithms,  
we only evaluate the performances for the reachability based variants including MFPT-VI, D-MFPT-VI, D-MFPT-VI-H1, D-MFPT-VI-H2.

Specifically, based on the reachability value computed for each state, we sort all states and partition the space into $n$ sub-spaces (partitions), using the partition heuristics of either D-MFPT-VI-H1 or D-MFPT-VI-H2.
In this way, states with similar reachability values are clustered together, e.g., in our simulation scenario the states of the highest reachability are grouped around the goal state. And, the smaller the reachability of the states, the farther they are from the goal state.
See Fig.~\ref{fig:plotsubstates} for an illustration.


\begin{figure}[t]
  \centering
  \subfigure[]
        {\label{fig:TP_DV}\includegraphics[height=1.3in]{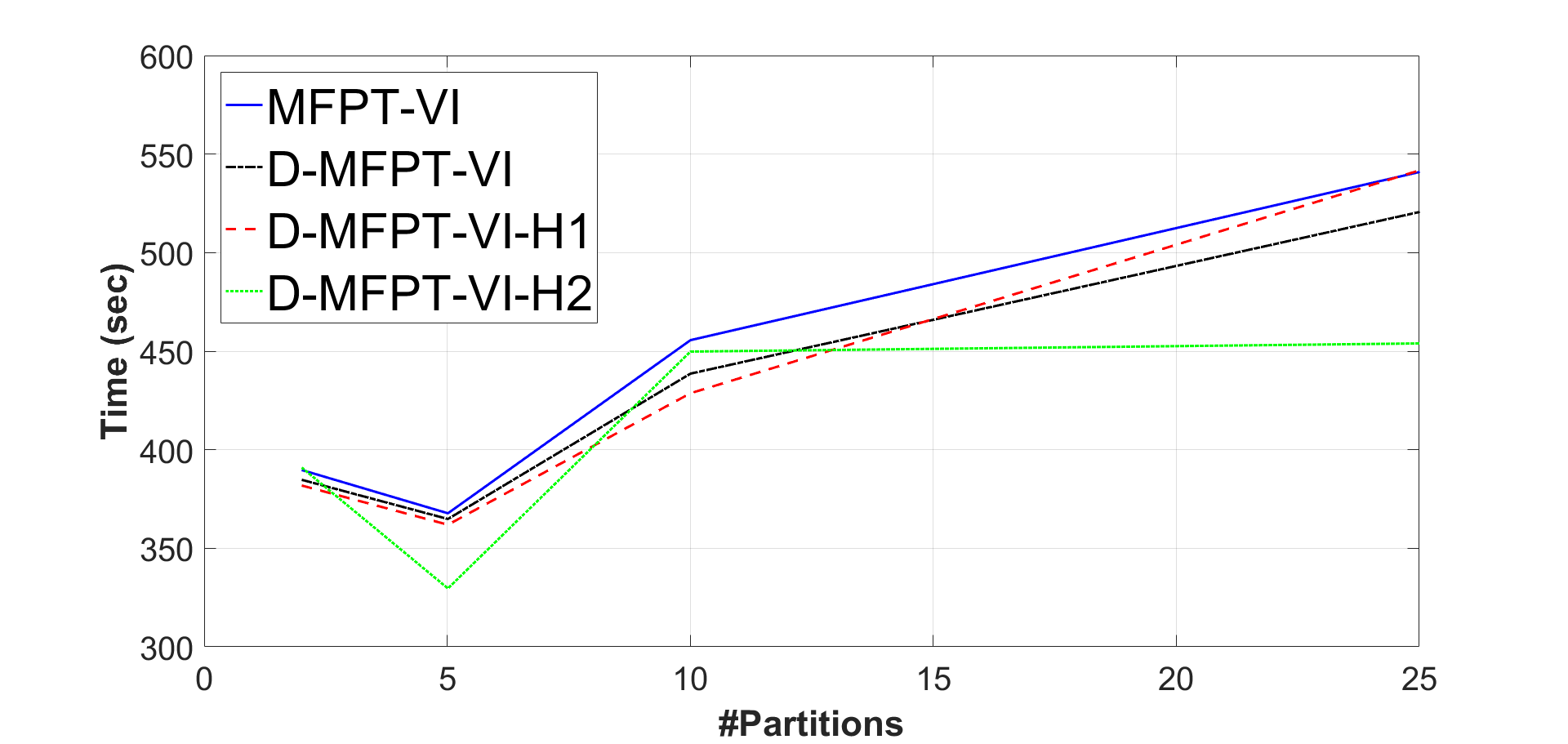}}
	\vspace{-10pt}
	\caption{\small Time comparisons across various variants under different size of partitions with the number of states set to 2500.
	}
	\label{fig:TimeVsPartitons}
	\vspace{-15pt}
\end{figure}

Fig.~\ref{fig:TimeVsPartitons} shows the time taken by the algorithms as the number of partitions $n$ changes. 
The results show that, given the same number of partitions, the D-MFPT-VI-H2 algorithm is in general the fastest among the three algorithms, 
indicating that partitioning the state space measured by states' potential impacts can achieve better performance in comparison to partitioning measured by the sizes of sub-spaces.
In contrast, the advantage of D-MFPT-VI-H1 is not obvious as it improves over  the basic MFPT-VI algorithm with a very small margin.

Note, although partitioning the space and exploring the partial space can accelerate the convergence, we have not decided the ideal number of partitions.
For example, in Fig.~\ref{fig:TimeVsPartitons} we can see that initially the time taken by the algorithms reduces as the number of partitions increases. However, after a certain point, the time taken begins to  increase as the number of partitions grows; we can also observe that all the three algorithms reach the minimal time almost simultaneously when the number of partitions is at around 5.
Finding the optimal partitioning number is difficult as the state (reachability) values can be very unstructured.  
Consequently, we use the experimental method to find such number.


\begin{figure}[t]
  \centering
        {\label{fig:3D_TSP1}\includegraphics[height=1.8in]{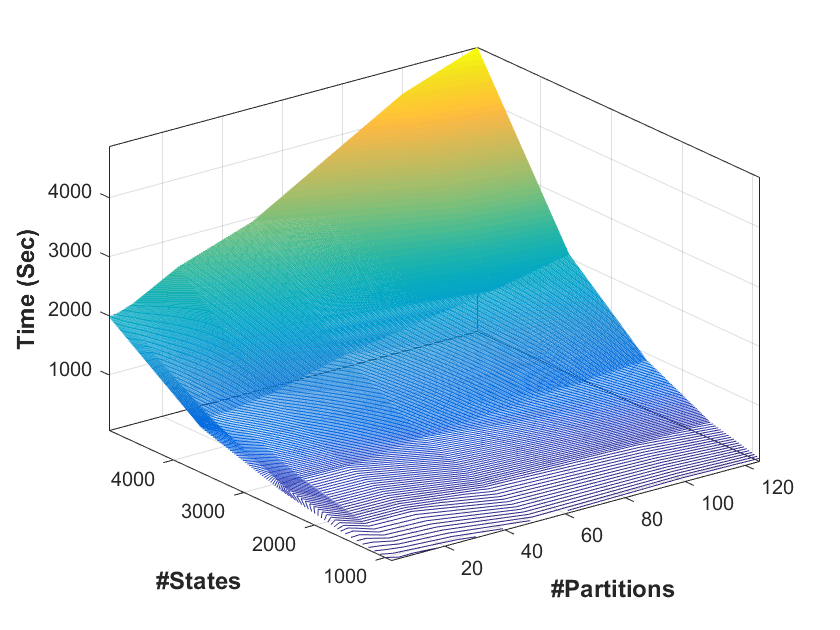}}
        \quad
	\vspace{-10pt}
	\caption{\small Time cost surface of D-MFPT-VI-H2 under different numbers of partitions and states. 
	}
	\label{fig:TimeVsPartitonsVSStates}
	\vspace{-15pt}
\end{figure}

We have plotted a clearer profile to capture the minimal-time partition number, as shown in Fig.~\ref{fig:TimeVsPartitonsVSStates}.
In the figure, the $x, y, z$ axes represent the number of states, number of partitions, and practical runtime, respectively. 
Note that, there is a ``crease" at the left side the 3D surface, and the bottom of the crease represent the optimal number of partitions in our heuristics.
Table ~\ref{min-time} shows the partition numbers that result in the minimum time taken to calculate the optimal policy for differing state-size. We can conclude that, the state space partition number need not be very large, and a partition number of around $2\sim5$ can result in near-optimal runtime.

\vspace{-5pt}
\begin{table}[]
\label{min-time}
\begin{tabular}{@{}l|lllll@{}}
\toprule
\toprule
 \textbf{State-Size} & \textbf{900} & \textbf{1600} & \textbf{2500} & \textbf{3600} & \textbf{4900} \\ \midrule
Partitions & 2 & 5 & 5 & 2 & 3 \\ \bottomrule \bottomrule
\end{tabular}
\centering
\vspace{3mm}
\caption{The number of partitions when the time taken by D-MFPT-VI-H2 is minimum to calculate the optimal policy across different state-size.}
\end{table}


\vspace{-35pt}
\section{Conclusions}
\vspace{-10pt}
\label{conclusion}

In this paper, we propose new heuristics for efficiently solving the MDPs.
Our proposed framework explores reachability of states using MFPT values, 
which characterizes the degree of difficulty of reaching given goal states. 
We then introduced the notion of backup differentials as an extension to the reachability characterization to capture more accurate impacts of states so that the prioritized sweeping can be ameliorated.
Also, we demonstrated that our reachability and differential based framework can be  further improved using partial-space sweeping strategies.
The experimental results show that in comparison with other state-of-the-art methods, our algorithms converge much faster, with less running time and fewer iterations.

\vspace{-20pt}
{
\bibliographystyle{abbrv}
\bibliography{reference}
}

\end{document}